\documentclass[letterpaper, 10 pt, conference]{ieeeconf}  

\IEEEoverridecommandlockouts                              

\usepackage{graphicx} 
\usepackage{mathptmx} 
\usepackage{times} 
\usepackage{amsmath} 
\usepackage{textcomp}
\usepackage{booktabs}
\usepackage{multirow}
\usepackage{balance}
\usepackage{soul,color}
\usepackage{comment}
\usepackage{array}
\usepackage{cite}
\usepackage{subcaption,booktabs}

\title{\LARGE \bf
Towards Interpretable Attention Networks for Cervical Cancer Analysis
}

\author{
Ruiqi Wang$^{1}$, 
Mohammad Ali Armin$^{1}$,
Simon Denman$^{2}$,
Lars Petersson $^{1}$, 
David Ahmedt-Aristizabal$^{1,2}$
\thanks{$^{1}$ CSIRO, DATA61, Canberra, Australia. 
{Corresponding author: \tt\footnotesize david.ahmedtaristizabal@data61.csiro.au}
}%
\thanks{$^{2}$ Image and Video Research Laboratory, SAIVT, Queensland University of Technology, Brisbane, Australia.%
}%
}

\begin{document}
\def\Ali#1{{\color{blue}{\bf [Ali:} {\it{#1}}{\bf ]}}}

\maketitle
\thispagestyle{empty}
\pagestyle{empty}

\begin{abstract}
Recent advances in deep learning have enabled the development of automated frameworks for analysing medical images and signals, including analysis of cervical cancer. Many previous works focus on the analysis of isolated cervical cells, or do not offer sufficient methods to explain and understand how the proposed models reach their classification decisions on multi-cell images.
%
Here, we evaluate various state-of-the-art deep learning models and attention-based frameworks for the classification of images of multiple cervical cells. As we aim to provide interpretable deep learning models to address this task, we also compare their explainability through the visualization of their gradients. 
We demonstrate the importance of using images that contain multiple cells over using isolated single-cell images. We show the effectiveness of the residual channel attention model for extracting important features from a group of cells, and demonstrate this model's efficiency for this classification task. 
This work highlights the benefits of channel attention mechanisms in analyzing multiple-cell images for potential relations and distributions within a group of cells. It also provides interpretable models to address the classification of cervical cells. 
\end{abstract}

\section{INTRODUCTION}

Cervical cancer is a serious health problem and it is one of the most common types of cancer in women worldwide\cite{canavan2000cervical}. With the development of promising computer vision techniques, more and more practical and efficient image analysis models exist to provide reliable auxiliary diagnosis results based on cell images. In cervical cell image classification tasks, the input can be an image showing a single isolated cell, or an image showing multiple cells. A model must classify the type of cell in the image (\textit{e.g} koilocytotic, metaplastic).

Various works have performed cervical cell classification using neural networks. 
Plissiti et al. \cite{plissiti2018sipakmed} applied VGG-16\cite{simonyan2014very}, a deep convolutional neural network, to classify isolated cervical cell images. 
Talo et al. \cite{talo2019diagnostic} proposed a DenseNet-161\cite{huang2017densely} model which improved upon the results of \cite{plissiti2018sipakmed}. 
To further improve performance, Haryanto et al.\cite{haryanto2020utilization} introduced a padding scheme to AlexNet\cite{krizhevsky2012imagenet}. 
Apart from deep neural networks, Win et al. \cite{win2020computer} combined various traditional machine learning methods such as random forests, support vector machines and k-nearest neighbors for segmentation and classification. 
GV et al. \cite{gv2019automatic} proposed a segmentation-free PCA based approach combined with a deep convolutional neural network to achieve the state-of-the-art result. 
However, these promising results are based on isolated cell images, and thus consider only a single isolated cell. Un-cropped cervical cell images contain multiple cells in different regions, and we refer to these as multi-cell images. Focusing only on a single cell discards vital information in the multi-cell image. For example, different types of cells in a multi-cell image have different distributions, and the relation between them varies. In addition, although these deep learning models achieve high performance for classifying isolated cells, they do not provide explainability and interpretabilty information for their model, which makes it difficult to understand the rationale behind their decisions. To enable adoption of these methods, we need improved explanations and interpretabilty to build user confidence and acceptance.

In this paper, we aim to develop interpretable deep learning models for the classification of multi-cell cervical cell images. Particularly, our work focuses on exploring the feasibility of adapting attention-based frameworks. Several prominent explainability methods for CNN-based models have been introduced including class activation mapping~\cite{zhou2016learning} and guided backpropagation~\cite{springenberg2014striving}.
While there is much interesting research within this field, it is immature and there are only a few works that investigate explanation methods for cervical cancer classification. We verify our models using an attribution prediction technique and compare the interpretable learning results offered by traditional CNN models and the proposed attention-based model.

Our main contributions are summarized as follows:
\begin{enumerate}
\vspace{-2pt}
\item We compare and introduce multiple deep learning models including a residual network, dense network, classic residual attention model and residual channel attention mechanisms for the purpose of classifying multi-cell cervical cell images.
\item We demonstrate the effectiveness of residual channel attention mechanisms for multi-cell cervical cell images, and show these to be robust and accurate. 
\end{enumerate}


\section{METHODOLOGY}

In this paper, we analyze different deep learning models for cervical cell classification and compare their performance. We introduce attention-based frameworks to extract important features from multi-cell cervical cell images. We aim to demonstrate the effect of the attention mechanism on the classification of multi-cell cervical images, which contain blank backgrounds with multiple cells of the same type. We also provide reliable explanations for how the attention mechanisms work, and show its interpretability through gradient visualization. 

\subsection{Traditional Deep Convolutional Neural Networks}
With the development of deep learning, various proposed deep convolutional neural networks have achieved success in classification, detection and segmentation tasks. We select the following two prominent traditional deep convolutional neural networks, which are already shown to be successful for cervical image classification\cite{talo2019diagnostic, hussain2020comprehensive}.

\subsubsection{Residual Convolutional Networks (ResNet)} 
Recent studies have shown the high performance of ResNet\cite{he2016deep}, which leverages the residual block structure, for image recognition and classification tasks. It uses residual (or skip) connections to allow information to more readily propagate through the network. ResNet models have many variants with different number of layers and different residual block structures. The baseline model in our experiments is based on ResNet50, which has 48 Convolutional layers, along with 1 MaxPool and 1 Average Pool layer.

\subsubsection{Dense Convolutional Networks (DenseNet)} DenseNet\cite{huang2017densely} simplifies the connectivity pattern between layers in a deep neural network, which significantly reduces the number of parameters and prevents learning from redundant feature maps. The structure of DenseNet is different from a traditional deep neural network, which concatenates the output feature maps of each layer with incoming feature maps, instead of summing them together. Similar to how ResNet is divided into residual blocks, DenseNet is divided into Dense Blocks with the same dimension as the feature maps, but with a different number of filters. Therefore, each layer in DenseNet has direct access to its preceding feature maps, and collect the new information it learned from the input. We chose one variant from this light and effective model, DenseNet101, for use in our experiments. 

\subsection{Attention-based frameworks}
Recently, attention mechanisms have shown an ability to achieve substantial improvements when added to deep learning models for various applications. Attention mechanisms helps a network to focus on important features in the data, and lead to more accurate decisions. Hence, we introduce two different attention mechanisms in our model for further experiments.

\subsubsection{Residual Attention Networks}
\label{subsec:ran}
The residual attention network\cite{wang2017residual} is a convolutional neural network based on the attention mechanism of~\cite{vaswani2017attention}. A naive attention learning creates a soft mask on the input to generate attention-aware features. However, it may lead to a performance drop as its dot product operation with a mask will degrade the value of features in deep layers. Residual attention networks solve this problem by using a trunk branch, a pre-activation ResNet unit~\cite{he2016identity} for feature extraction; and a mask branch, which uses a bottom-up top-down structure for learning the mask. In the residual attention network, it adds the attention module after each residual layer, making features clearer as depth increases. We use this residual attention network based on the ResNet50 framework as our first attention-based model. 

\subsubsection{Residual Channel Attention}
For image super-resolution, it is necessary to avoid learning abundant low-frequency information from low-resolution inputs and features. These unnecessary features are treated equally across channels. The residual channel attention network (RCAN)\cite{zhang2018image} is made up of residual-in-residual (RIR) blocks with channel attention, which ignores abundant low-frequency information through skip connections within blocks, and adaptively rescales features between channels. Channel attention combines channel-wise features with different weights by exploiting the interdependencies among them. Therefore, for highly accurate image super-resolution, it is more flexible and more powerful for feature extraction between channels. We aim to use this RCAN model to verify the effectiveness of attention mechanisms for use with multi-cell images, and compare it with the residual attention networks (see Section \ref{subsec:ran}). 
 
\subsection{Explainability and interpretability}
Lack of transparency is identified as one of the main barriers for AI adoption in clinical practice. A step towards making AI trustworthy is the development of explainable AI methods. 
In this analysis, we adopted the integrated gradient model\cite{sundararajan2017axiomatic} to illustrate the advantages of these methods for helping clinicians to identify the parts of the images that are critical in the decision.

The \textit{integrated gradient model} computes the attribution of the prediction of the deep neural network by using the gradient operation. Different to previous attribution techniques, it is characterized by two axioms: sensitivity and implementation invariance; which makes it more flexible and easier to apply to a variety of deep networks.

\begin{figure*}[htbp!]
\centering
    \subfloat[\centering Superficial]{{\includegraphics[width=3cm]{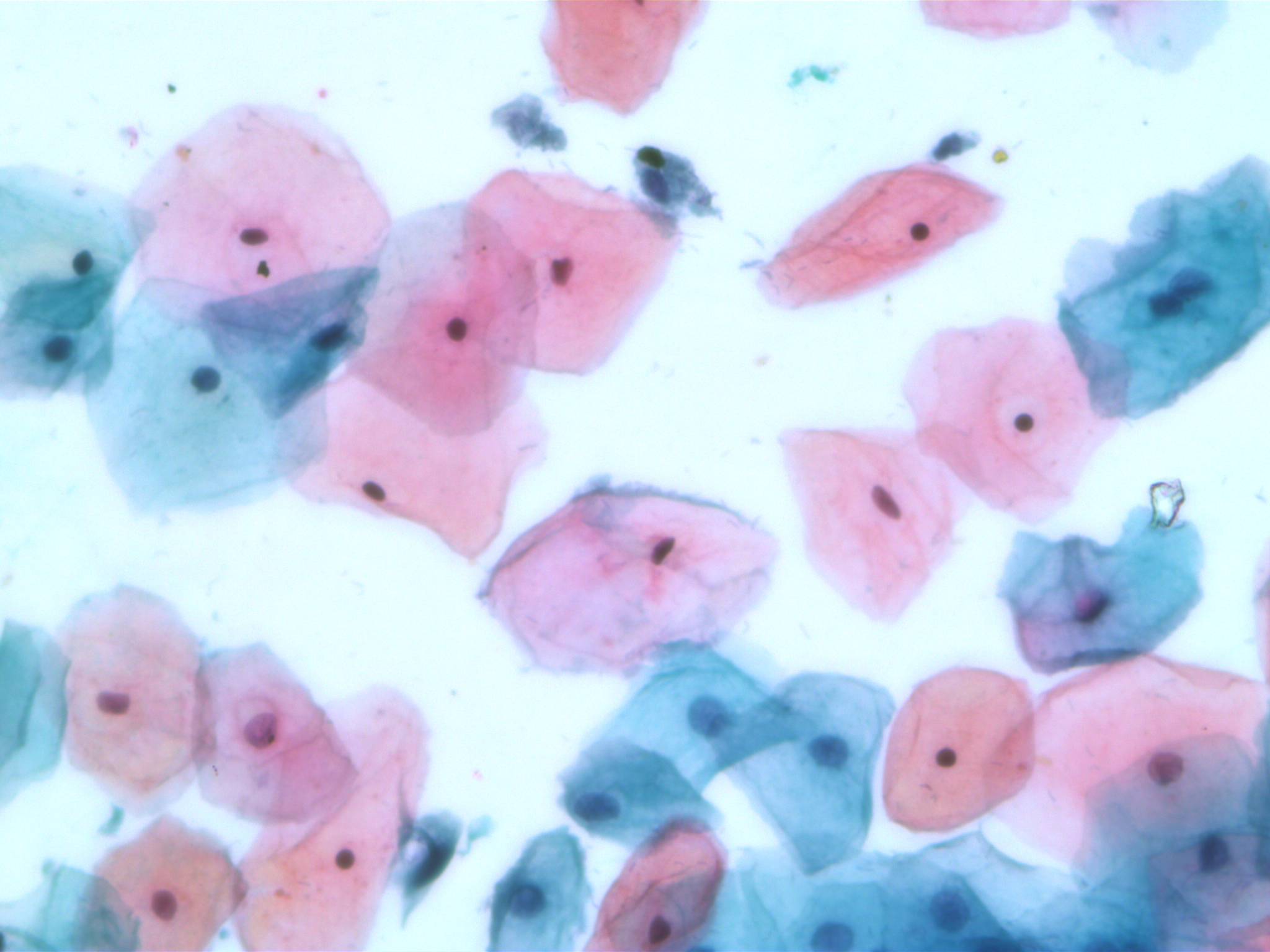} }}%
    \hspace{1mm}
    \subfloat[\centering Parabasal]{{\includegraphics[width=3cm]{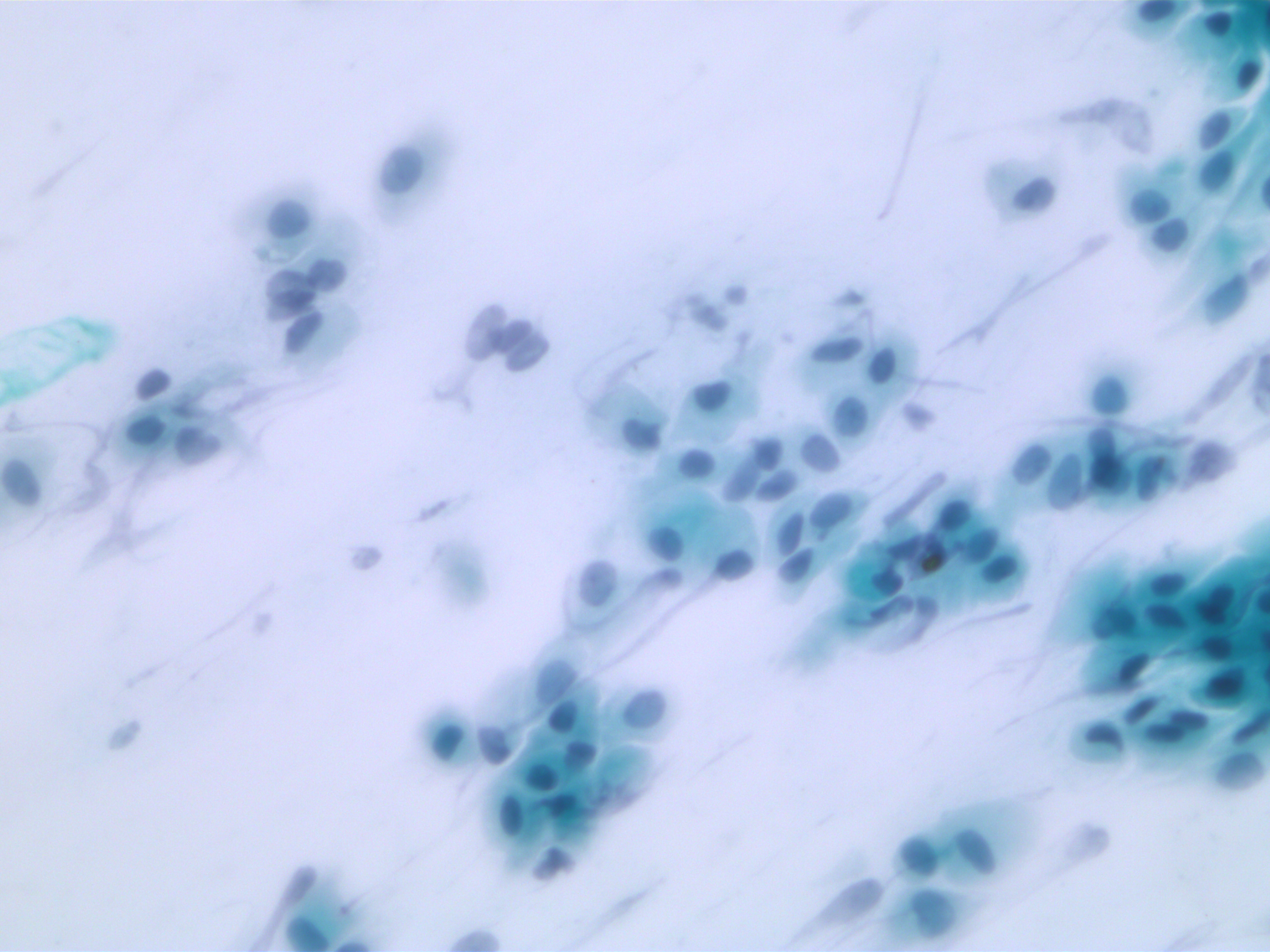} }}%
    \hspace{1mm}
    \subfloat[\centering Koilocytoti]{{\includegraphics[width=3cm]{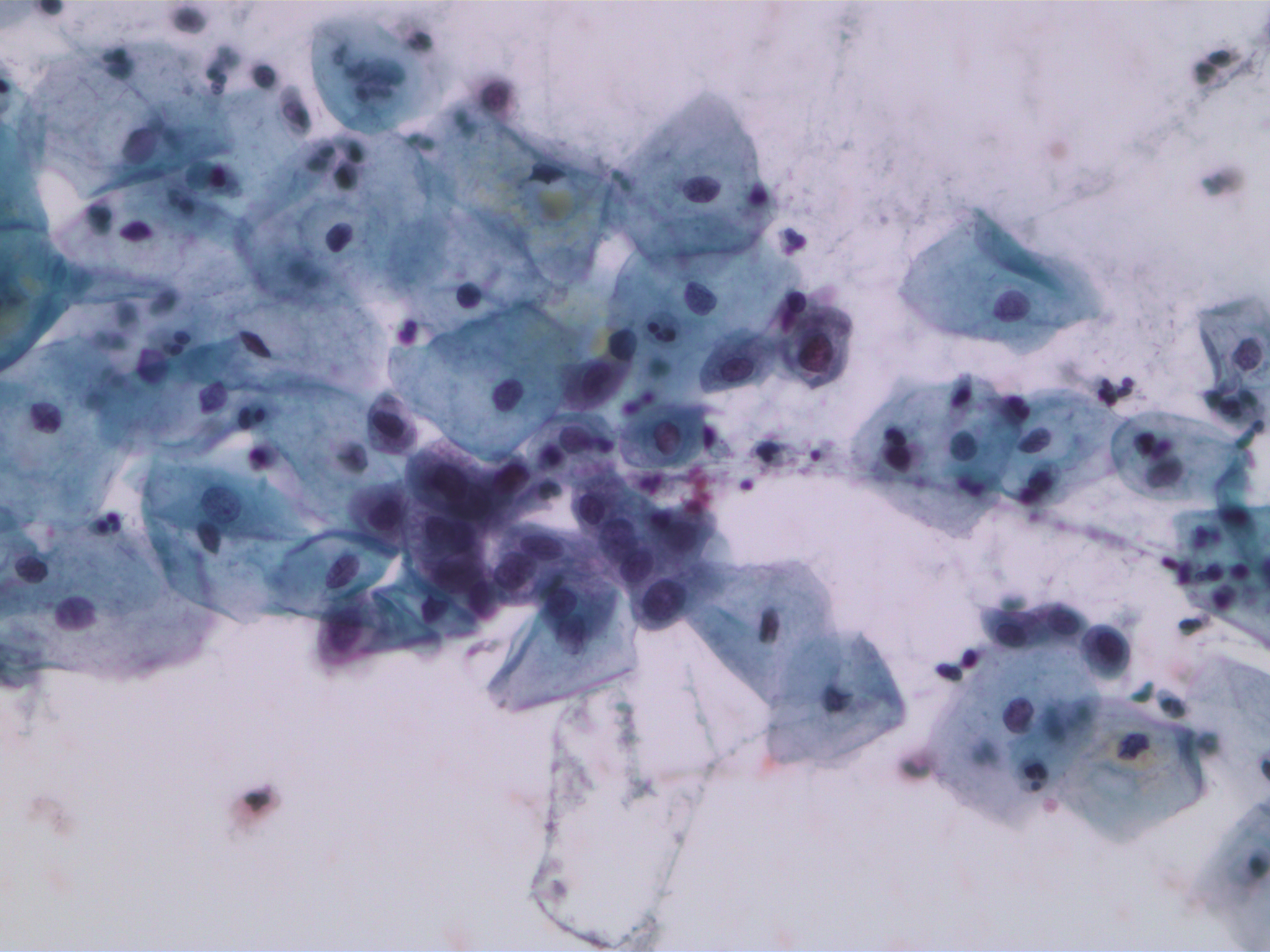} }}%
    \hspace{1mm}
    \subfloat[\centering Dyskeratoti]{{\includegraphics[width=3cm]{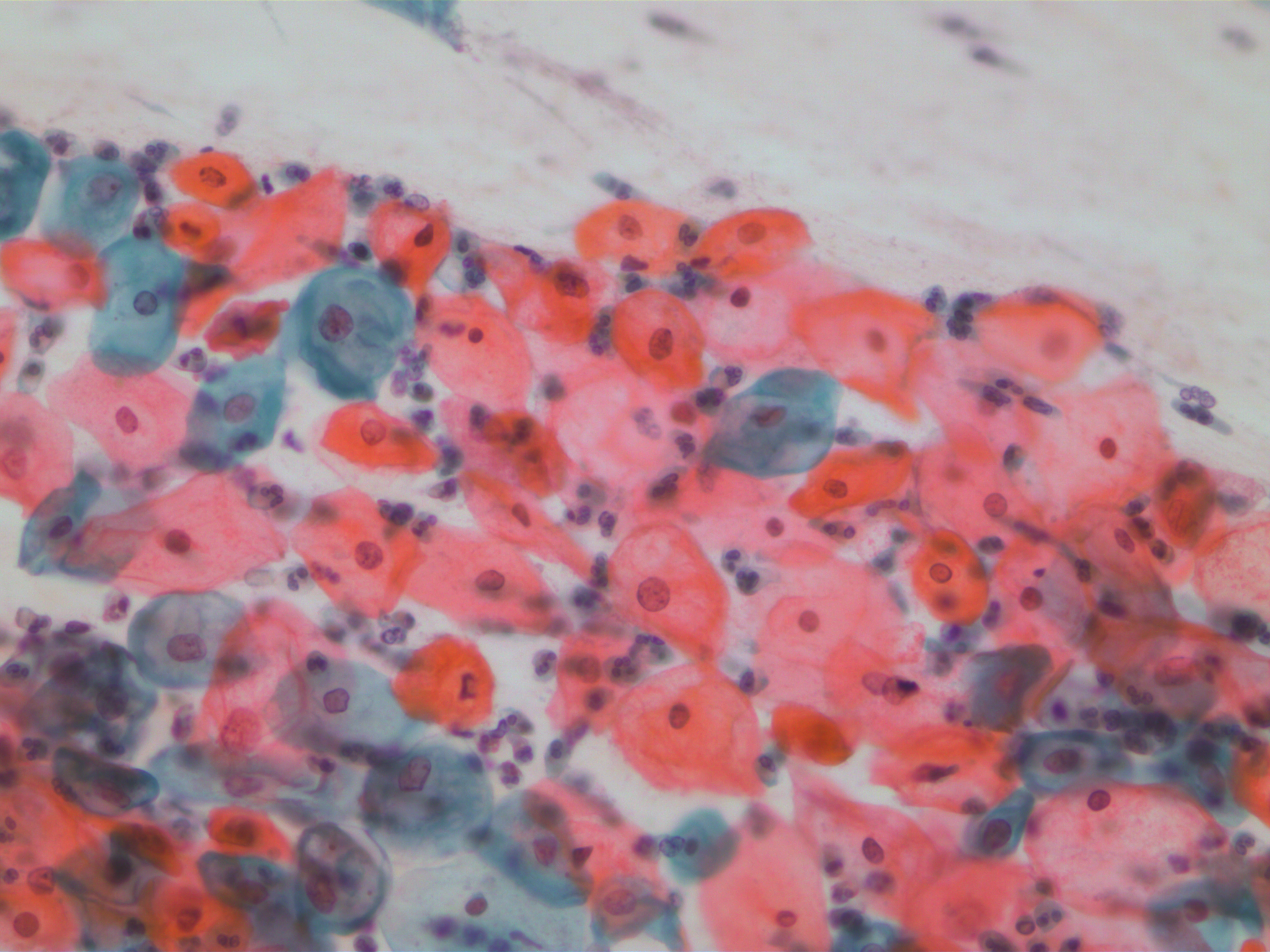} }}%
    \hspace{1mm}
    \subfloat[\centering Metaplastic]{{\includegraphics[width=3cm]{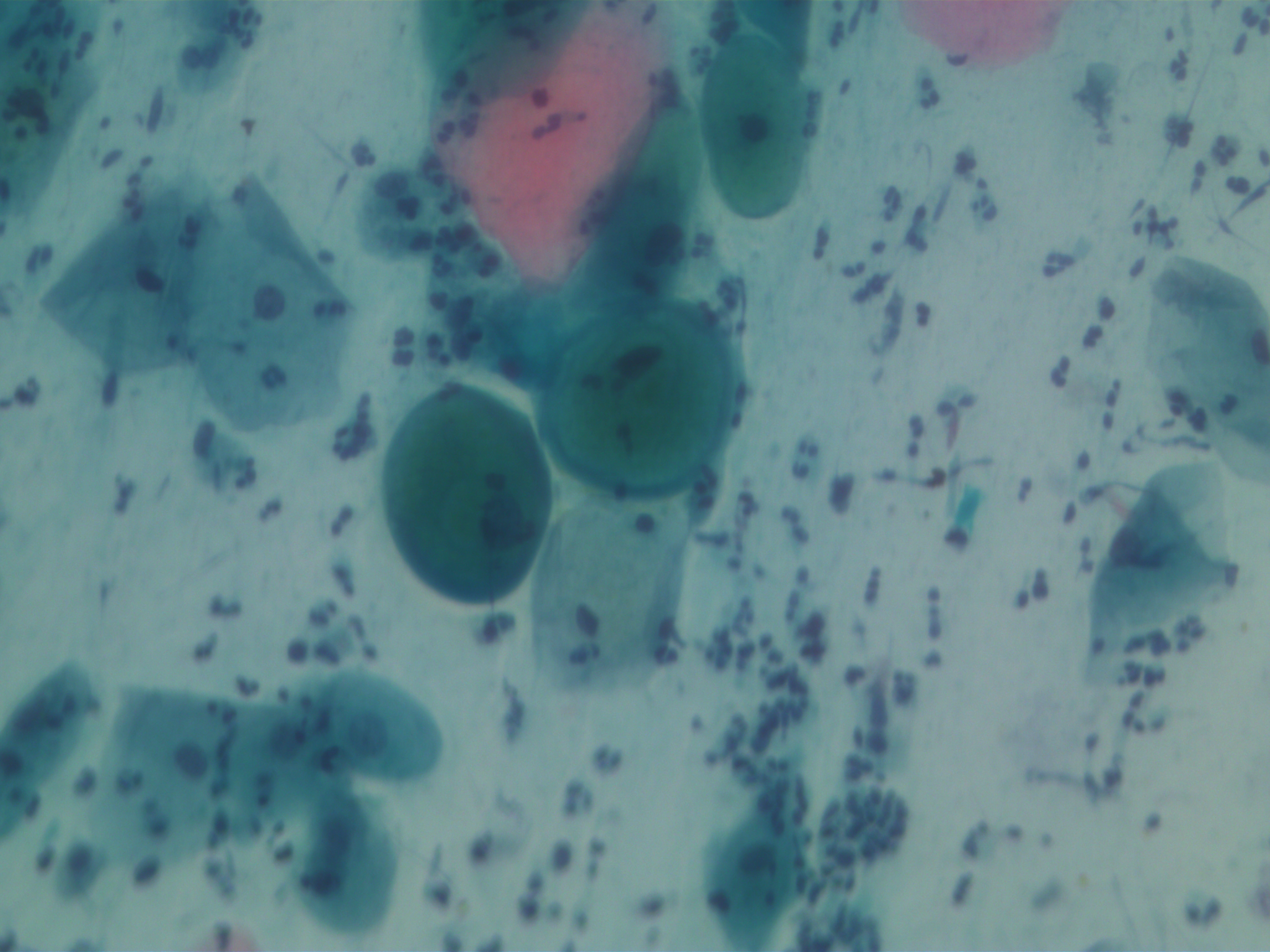} }}%
\vspace{-3pt}
   \caption{Multi-cell images of the five categories considered.
   }
\vspace{-6pt}
\label{fig:data}
\end{figure*}

\section{EVALUATION}

\subsection{Datasets and experimental setup}
The performance of each proposed method is evaluated on the SIPaKMeD dataset\cite{plissiti2018sipakmed}, which is an open-source cervical cell image database. It consists of two different types of data: 966 multi-cell images and 4049 isolated cell images. There are five different categories of cervical cells in this dataset: superficial-intermediate, parabasal, koilocytotic, dysketarotic and metaplastic. Sample images from each class are shown in Fig.~\ref{fig:data}. Class distribution details are presented in Table~\ref{table:data}. These cell images were acquired using a high resolution CCD camera connected to an optical microscope. From the data distribution in Table~\ref{table:data} it can be noted that the classification task for multi-cell images is more challenging as, apart from the imbalanced class distribution, the volume of data is smaller than for the isolated cell images. 
\begin{table}[b!]
\centering
\caption{Data distribution of the SIPaKMeD dataset}
\label{table:data}
\resizebox{0.35\textwidth}{!}{%
\begin{tabular}{ccc}
\hline
Category     & Multi-cell Images & Isolated Images \\ \hline
Dyskeratotic & 223         & 813             \\
Koilocytotic & 238         & 825             \\
Metaplastic  & 271         & 793             \\
Parabasal    & 108         & 787             \\
Superficial  & 126         & 813             \\ \hline
\end{tabular}%
}
\end{table}
In our experiments, we focus on multi-cell images only, as isolated cell images contains features of a single cell only and thus do not contain information on cell distributions and relationships between cells. We also aim to demonstrate that some of these isolated cell images may suffer from the lack of cell information. 
To verify the capability and effectiveness of each model, we also applied them to the isolated cell images. Both types of data (multi-cell and single-cell images) are split such that 70\% of the data samples are allocated for training, 20\% for validation, and 10\% for testing.

\vspace{-4pt}
\subsection{Evaluation metric and implementation}
As this is a multi-class classification task with an uneven class distribution, we use a weighted-F1 score to measure performance. This is calculated as follows,
\begin{equation}
\text{Weighted} \; F_1 = \sum_{i=1}^{5} \dfrac{2 \times \text{precision}_i \times \text{recall}_i} {\text{precision}_i + \text{recall}_i} \times w_i,
\end{equation}
where $w_i$ is the weight of the $i-th$ class and depends on the number of positive examples in that class.

The overall accuracy on the test set is also an evaluation metric, which simply shows the overall performance of different models. Categorical cross-entropy loss and the Adam optimizer~\cite{kingma2014adam} (learning rate=$0.001$, other parameters are default) are used to train the models. Models are trained for 50 epochs with a mini-batch size of 16, as we found more epochs lead to overfitting. All models were implemented in Pytorch\cite{paszke2019pytorch}.

\vspace{-4pt}
\subsection{Experimental results and discussion}

An evaluation of all proposed models and baseline methods for each dataset are shown in Table~\ref{table:results}. 

From the accuracy results on both isolated and multi-cell images, it is obvious that the DenseNet-121 model with the residual channel attention mechanism has a significant advantage compared to others. The original DenseNet-121 model achieves better results than the ResNet50, as it is deeper and able to extract more hidden features. The attention-based model of DenseNet121 also has the highest accuracy on the test set.

However, it is worth noting that the residual attention model based on ResNet-50 decreases the accuracy of ResNet-50. To explore its prediction results in more details, we compute the F1-score for each model on each class, shown in the lower section of Table~\ref{table:results}. The baseline ResNet-50 model has difficulty classifying koilocytotic and metaplastic cells correctly since both are in large size and some koilocytotic cells are a type of metaplastic cell; although there are still slight differences between them in their color, contour, size and shape. Therefore, the introduction of attention in ResNet-50 targets these differences between these two classes and improves the precision of them. However, as in the residual attention model, the attention layer is added after each residual layer and it learns similar noise for other classes. Therefore, although it improves upon the weakness of the baseline model, it decreases the precision of other well-performing classes and thus lowers the overall accuracy. 

The performance of residual channel attention DenseNet-121 is outstanding in four of the five classes. Comparing with the original DenseNet-121 model, the introduction of residual channel attention significantly improves the precision on the dyskeratotic class and koilocytotic class with a slight loss in precision for metaplastic cells. This result also gives strong proof of the effectiveness of the channel attention mechanism in this cervical cell classification task. 

\begin{figure}[b!]
\centering
\begin{subfigure}[b]{0.45\textwidth}
   \includegraphics[width=1\linewidth]{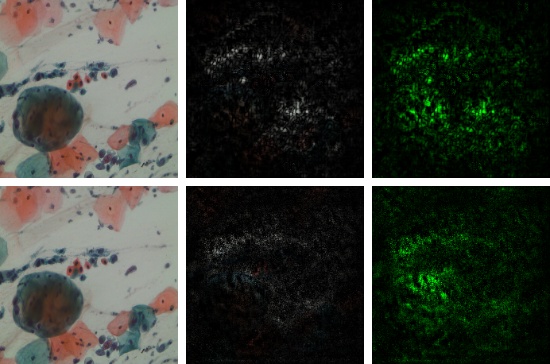}
   \caption{DenseNet-121 without Attention}
   \label{fig:Ng1} 
\end{subfigure}

\vspace{3mm}
\begin{subfigure}[b]{0.45\textwidth}
   \includegraphics[width=1\linewidth]{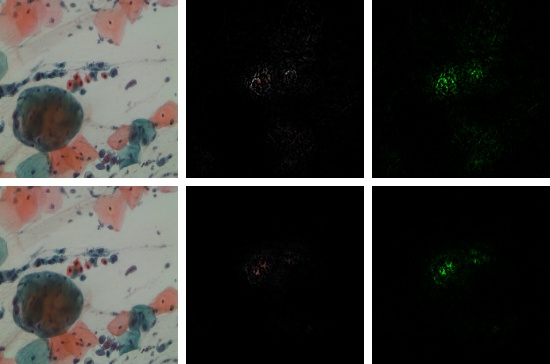}
   \caption{DenseNet-121 with Residual Channel Attention}
   \label{fig:Ng2}
\end{subfigure}
\caption{Attribution Prediction Gradient Visualization. Columns from left to right: the original image, gradient and integrated gradient overlay, gradient and integrated gradient.}
\label{figure:gradient}
\vspace{-8pt}
\end{figure}

In order to explore more details of the attention learning progress, we visualize the gradient attribution prediction of DenseNet-121 and the residual channel attention based model in Fig.~\ref{figure:gradient}.

\begin{table}[]
\centering
\captionsetup{justification=centering}
\caption{Results}
\label{table:results}
\begin{subtable}[]{0.485\textwidth}
    \resizebox{\textwidth}{!}{%
    \begin{tabular}{@{}ccccc@{}}
    \toprule
                   & ResNet-50 & DenseNet-121\cite{talo2019diagnostic} & RAN-ResNet-50 & RCAN-DenseNet-121 \\ \midrule
    Accuracy & 95.11\%   &  95.84\%      & 94.13\%       & \textbf{96.33\%}  \\ \bottomrule
    \end{tabular}%
    }
    \caption{Overall accuracy on isolated cervical cell images}
\end{subtable}

\begin{subtable}[]{0.485\textwidth}
    \resizebox{\textwidth}{!}{%
    \begin{tabular}{@{}ccccc@{}}
    \toprule
                   & ResNet-50 & DenseNet-121\cite{talo2019diagnostic} & RAN-ResNet-50 & RCAN-DenseNet-121 \\ \midrule
    Accuracy & 85.15\%   & 89.11\%      & 84.16\%       & \textbf{91.09\%}  \\ \bottomrule
    \end{tabular}%
    }
    \caption{Overall accuracy on multi-cell cervical cell images}
\end{subtable}

\begin{subtable}[]{0.485\textwidth}
    \resizebox{\textwidth}{!}{%
    \begin{tabular}{@{}ccccc@{}}
    \toprule
                 & ResNet-50 & DenseNet-121   & RAN-ResNet-50 & RCAN-DenseNet-121 \\ \midrule
    Dyskeratotic & 0.869     & 0.898          & 0.851         & \textbf{0.978}    \\
    Koilocytotic & 0.744     & 0.809          & 0.783         & \textbf{0.869}    \\
    Metaplastic  & 0.839     & \textbf{0.929} & 0.877         & 0.896             \\
    Parabasal    & 0.96      & 0.957          & 0.88          & \textbf{1.0}      \\
    Superficial  & 0.846     & \textbf{0.889} & 0.815         & \textbf{0.889}    \\ \bottomrule
    \end{tabular}%
    }
    \caption{F1 score of each class on multi-cell cervical cell images}
\end{subtable}
\vspace{-14pt}
\end{table}

From the gradient visualization results of a random test image from the dyskeratotic category, we can observe that DenseNet-121 considers features from the input image from a large area of cells, including a part of the background. Conversely, the model with residual channel attention is more focused on a small specific region of the multi-cell image, which means it makes the decision from a few cells only, ignoring the background and other noisy information. From this interpretable visualization result of the deep neural network, we also understand how the attention mechanism works in classifying different cervical cells. It gives more weight in parts of the cell groups for classification, focusing on parts which contains useful features and relations. This visualization results also provide information about the specific regions in the multi-cell images, which helps to assist the identification of cervical cells of particular interest for experts in a real world application.

Our current evaluation results show that the residual channel attention mechanism is efficient for analyzing the multi-cell cervical cell images and classifies them more precisely. It also offers information regarding hidden relations between cell groups in the multi-cell image, and it is worth noting that the attention often falls on a specific group of cells.
There may also be other factors such as the distribution of different classes of cervical cells which can be analyzed. 
%
CNNs have been commonly used in the digital pathology domain for the classification of fixed-sized biopsy image patches. However, learning over patch-wise features limits the model to capturing global contextual information.
Recently, graph data representations have attracted significant attention in the analysis of histological images~\cite{aygunecs2020graph} due to their ability to represent the tissue architecture by modeling a tissue section as a multi-attributed spatial graph of its constituent cells. Graph-based representations can encode the spatial relationships across the patches for fine-grained classification~\cite{aygunecs2020graph}. In future work, relations between cells in multi-cell images can be analyzed through this technique. 

\section{CONCLUSIONS}

We introduce and compare deep convolutional neural networks with different attention mechanisms for cervical cell classification. We also give a detailed explanation of how interpretation methods can be applied to classification results for cervical cell images. Our experiments and analysis show that the residual channel attention framework is effective in distinguishing between features for different classes and isolating a specific region of interest for multi-cell cervical cell images.

\balance


\bibliographystyle{IEEEtran}
\bibliography{refs}

\vspace{0.5cm}

\end{document}